\documentclass[a4paper,twoside]{article}

\usepackage{epsfig}
\usepackage{subcaption}
\usepackage{calc}
\usepackage{amssymb}
\usepackage{amstext}
\usepackage{amsmath}
\usepackage{amsthm}
\usepackage{multicol}
\usepackage{pslatex}
\usepackage{apalike}
\usepackage{algorithm2e}
\usepackage[bottom]{footmisc}
\usepackage{SCITEPRESS} 
\usepackage{url}
\usepackage{cuted}
\usepackage{multirow}

\begin{document}

\title{Benchmarking ResNet Backbones in RT-DETR: Impact of Depth and Regularization under environmental conditions}

\author{\authorname{Pamela Barboza\sup{1}\orcidAuthor{0009-0000-5299-4252}, Víctor Castelli\sup{1}\orcidAuthor{0009-0006-2917-6343},
 Belén Pereira \sup{1}\orcidAuthor{0009-0004-2627-8630},
 Ricardo Grando \sup{1}\orcidAuthor{0000-0002-2939-5304},
Bruna de Vargas \sup{1}\orcidAuthor{0000-0001-6044-0671},
Augusto Calfani \sup{1}\orcidAuthor{0009-0001-8163-208X}, Gabriela Flores\sup{1}\orcidAuthor{0009-0000-2293-5702}}
\affiliation{\sup{1}, Robotics and Artificial Intelligence Laboratory, Technological University of Uruguay, Rivera, Uruguay}
\email{\{pamela.barboza, victor.castelli, ricardo.bedin, bruna.devargas\}@utec.edu.uy, \{augusto.calfani, belen.pereira, luisa.flores\}@estudiantes.utec.edu.uy}
}


 \abstract{Visual perception plays a central role in competitive robotics, where environmental variations can directly affect real-time detection performance. The related literature on transformer-based detectors lack information regarding the impact of backbone scale and environmental settings on model performance.  This work presents a comparative evaluation of RT-DETR for detecting round objects under environmental and hyperparameter variations relevant to competitive robotics. Four ResNet backbones (ResNet18, ResNet34, ResNet50, and ResNet101) were compared using dropout rates, analyzing their effect on confidence and accuracy. All models were trained under the same configuration and evaluated under changes in lighting and background contrast. Environmental conditions primarily impact prediction confidence, while inference latency remains largely unaffected and classification accuracy stays consistently high, approaching or above 1.00 in most cases. Two distinct behaviors were observed. Under illumination variation, ResNet50 achieves the best trade-off, combining near-perfect accuracy, confidence values up to approximately 0.869 and latency around 0.058–0.059 ms. Under background variation, ResNet34 provides the most balanced performance, reaching near-perfect accuracy and higher confidence values up to approximately 0.887. These results indicate that the optimal architecture depends on the type of environmental variation, with intermediate-depth models offering the best balance between performance and efficiency.}


\onecolumn \maketitle \normalsize \setcounter{footnote}{0} \vfill

\section{\uppercase{Introduction}}
\label{sec:introduction}

\noindent In recent years, real time object detection has become a core component of autonomous robotic systems. This need is more clear in competitive environments, where robots must detect and track small objects under strict temporal constraints. In robot sports and service competitions, visual perception is often required to localize round or nearly round objects such as balls, markers, or game elements while the scene changes continuously. In scenarios such as robot soccer, basketball inspired tasks, and other competition settings, object detection must remain reliable despite motion, viewpoint changes, limited processing time, and environmental variability.

Among these challenges, illumination and background conditions remain two of the most critical factors. In the RoboCup Small Size League, Weinmann reports that the currently used SSL-Vision framework is strongly limited by variable lighting conditions, including uneven spatial illumination, temporal illumination changes, natural sunlight, and line frequency flickering, all of which can cause color misclassification and force manual recalibration during operation \cite{weinmann2024}. The same study also shows that the problem is not restricted to brightness alone. Background and scene composition directly affect color based perception, since gray carpets prevent dominant hue based field estimation, wooden walls reduce the usefulness of brightness cues, and several competition datasets present reduced color contrast, blue shifted white balance, oversaturated colors, and specular glare \cite{weinmann2024}.

A related difficulty is also observed in humanoid competition environments. In a FIRA HuroCup obstacle detection study, object perception was explicitly evaluated under dark and bright illumination using different color representations. The authors reported that the HSV Value channel outperformed the RGB Blue channel in both low light and bright conditions, reaching 71.43\% versus 50\% in dark scenes and 82.86\% versus 61.42\% in bright scenes, which confirms that detection performance is strongly affected by illumination and by the selected color space \cite{prabowo2024}. 

These limitations are particularly relevant in competition scenarios that depend on the reliable detection of small round objects. In RoboCup SSL, for example, the orange ball is intentionally selected to maximize contrast against the green field, while additional colored blobs are used to identify teams and robot orientation \cite{weinmann2024}. This illustrates a broader design principle in robotic competitions: object perception often depends on color contrast assumptions that may not hold under adverse illumination or under specific backgrounds. When these assumptions break, detector confidence and classification stability may degrade even if the object geometry remains simple.

In parallel with these environmental factors, the performance of transformer based detectors also depends on the feature extractor used in the vision pipeline. Deep convolutional neural network architectures such as the one presented in YOLO (You Only Look Once) have dominated real time detection due to their favorable trade off between accuracy and speed \cite{Redmon2016}. More recently, Transformers have reshaped computer vision by enabling end-to-end object detection without non-maximum suppression, as demonstrated by RT-DETR \cite{Carion2020,Vaswani2017}. In this context, RT DETR was introduced as a real time adaptation that combines a hybrid encoder with efficient attention mechanisms to reduce latency while preserving competitive accuracy \cite{Zhao2023}.

\begin{figure}[ht]
  \centering
  \includegraphics[width=1\linewidth]{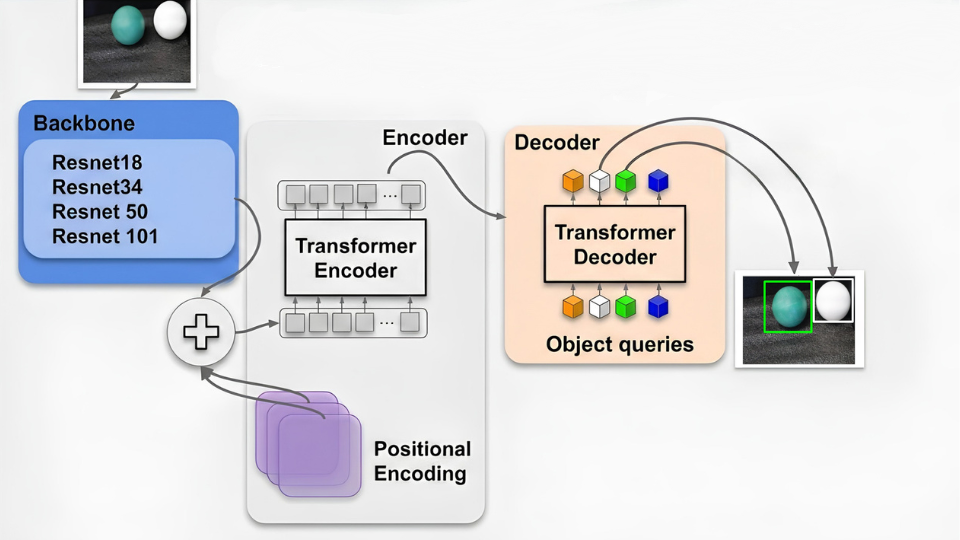}
  \caption{Schematic representation of the experimental setup.}
  \label{fig:Intro}
\end{figure}

In this context, this work focuses on comparing different backbones in RT-DETR to evaluate performance in different scenarios that can be found in the competition of robotics. In Figure \ref{fig:Intro}  shows the schematics of the experiments.

The main contributions of this work are summarized as follows:
\begin{itemize}
    \item A systematic evaluation of RT-DETR with backbone architectures of different depth, from ResNet18 to ResNet101, in order to analyze how model scaling affects detection accuracy, confidence, and inference latency.
    
    \item A controlled experimental analysis of two relevant sources of environmental variability, namely illumination changes and background contrast changes, which are common in robotic competition settings.
    
    \item An assessment of the effect of dropout regularization on RT-DETR performance, considering both detection reliability and temporal stability under different operating conditions.

\end{itemize}

The remainder of the paper is organized as follows. Section 2 reviews related work on real-time object detection and transformer-based architectures. Section 3 presents the methodology, including the dataset description, network configurations, and experimental setup. Section 4 reports the experimental results, section 5 analyzing the impact of backbone depth and dropout regularization across different lighting and background conditions. Section 6 concludes the paper.


\section{\uppercase{RELATED WORK}}
In recent years, object detection has shifted from region proposal-based methods to fully end-to-end architectures optimized for real-time performance. Two stage detectors, such as Faster R-CNN \cite{Redmon2016}, favored accuracy by explicitly generating regions of interest, but at a high computational cost. Single stage detectors such as YOLO \cite{Redmon2016} and its subsequent versions \cite{yolov5,yolov8,Wang2023}, as well as SSD \cite{Liu2016}, integrated localization and classification into a single regression task, significantly improving inference speed.
However, these approaches typically rely on Non Maximum Suppression (NMS) as a post-processing step, which introduces handcrafted confidence and IoU thresholds and can delay end-to-end inference in real-time systems \cite{zhao2024detrs}.

The introduction of Transformers transformed object detection. DETR \cite{Carion2020} reformulated detection as a direct set prediction problem using self attention mechanisms and a bipartite matching loss that removes the need for anchors and Non Maximum Suppression. Despite its strong theoretical foundation, DETR suffered from slow convergence and high computational complexity. To address these limitations, variants such as Deformable DETR \cite{Zhassuzak2025} introduced multi scale deformable attention to accelerate training and improve efficiency.

RT DETR \cite{Zhao2023} introduced an architecture optimized for real time performance with a hybrid CNN Transformer encoder and efficient multi scale feature fusion strategies. These optimizations reduce latency while maintaining the end-to-end framework. Subsequent versions have improved stability and training procedures without altering the overall design. These advances position transformer based detectors as alternatives to fully convolutional architectures in embedded and robotic applications.

Backbone selection remains a structural design decision. The ResNet family \cite{He2016} introduced residual connections that enable the training of deep networks without degradation. Shallower variants such as ResNet18 and ResNet34 reduce computational cost and latency, whereas deeper models such as ResNet50 and ResNet101 increase representational capacity at the expense of inference time \cite{wang2025rt}. While greater depth generally improves global metrics on complex datasets, several studies report diminishing returns in controlled environments or tasks involving geometrically simple objects \cite{ELHARROUSS2024100645}

Environmental conditions also affect detector robustness. Classical studies in digital image processing have shown that illumination changes alter contrast, color distribution, and signal to noise ratio \cite{Gonzalez2008,Szeliski2010}. Recent research on detection under low light conditions demonstrates that brightness and contrast degradation reduce object background separability, harming deep learning based detectors \cite{Loh2019,Chen2018}. In addition, controlled robustness evaluations of deep neural networks show that common perturbations such as brightness variations and shadows lead to measurable drops in accuracy \cite{Hendrycks2019}. These findings have motivated the use of normalization techniques and data augmentation to mitigate such effects.

Regarding regularization, Dropout was introduced as a method to reduce overfitting by randomly deactivating neurons during training \cite{Srivastava2014}. Although its effectiveness is well established in conventional deep networks, its interaction with with widely used techniques such as Batch Normalization \cite{Ioffe2015} and contemporary attention-based architectures has not been fully explored in real time transformer detectors.

Although the literature has extensively compared convolutional and transformer based detectors \cite{Carion2020,zhao2024detrs,zhu2021}, there is limited combined analysis of how backbone depth, Dropout regularization, and illumination conditions affect RT DETR architectures. This work studies this interaction through a controlled experimental evaluation, measuring the latency confidence trade off in the detection of simple geometric objects under different lighting and background conditions.

\section{\uppercase{Methodology}}

\noindent In this study, the performance of the RT-DETR architecture is evaluated using different ResNet backbones in order to determine the optimal balance between detection accuracy and computational efficiency.

\subsection{Network Architectures and Training}

To systematically assess the impact of feature extraction capacity on detection performance, the RT-DETR architecture was implemented using four different variants of the ResNet family as backbones: ResNet18, ResNet34, ResNet50, and ResNet101. This selection enables a direct evaluation of the trade-off between representational depth, detection accuracy, and computational latency. 

Furthermore, to evaluate the effect of stochastic regularization on model confidence and its ability to generalize under varying environmental conditions, each backbone configuration was trained under two distinct modes:

\begin{itemize}
    \item \textbf{Standard Mode:} No regularization applied (Dropout rate of 0.0).
    \item \textbf{Regularized Mode:} A dropout rate of 0.2 applied to the attention mechanisms of the transformer encoder.
\end{itemize}

\subsection{Dataset Description and Preprocessing}

The dataset employed in this work aggregates multiple publicly available repositories hosted on the Roboflow Universe platform \cite{roboboccia,pingpong_market5,tabletennis_hupo,yellowball_nzct5,yellowgolf,buoys_dataset}. In total, the dataset contains 8,806 images featuring a diverse collection of spherical objects. The focus on spherical target objects (specifically balls of various colors and diameters) was intentionally chosen to mirror the visual constraints of competitive robotics, such as the RoboCup or HuroCup leagues. In these environments, robots must reliably track geometrically simple, color-coded elements under dynamic lighting; thus, evaluating backbones strictly on these simple geometries isolates the network's capacity to handle color and illumination variations without the confounding variable of complex object shapes.

Originally, the dataset comprised five object classes: \textit{blue\_ball}, \textit{green\_ball}, \textit{orange\_ball}, \textit{red\_ball}, and \textit{white\_ball}. However, preliminary inspection revealed a strong chromatic similarity between the red and orange classes, particularly under artificial and reduced lighting conditions. To prevent ambiguity in class boundaries and inter-class confusion caused by overlapping hue distributions, the \textit{red\_ball} class was explicitly excluded. Consequently, all experiments were conducted using four clearly defined categories: blue, green, orange, and white balls. 

These images were captured under controlled experimental setups that preserve intentional variations in illumination (artificial and reduced lighting), background color (white and black surfaces), and object positioning. All images were manually annotated with bounding boxes and exported in YOLO format to ensure compatibility with the selected detection architectures. 

\textbf{Dataset Split.} The dataset was partitioned into training (7,549 images, approx. 85\%), validation (852 images, approx. 10\%), and test (405 images, approx. 5\%) subsets. This division ensures sufficient data for model learning while preserving independent subsets for unbiased performance evaluation.

\textbf{Preprocessing and Static Augmentation.} To prepare the data for training, all images were automatically oriented using Roboflow’s Auto-Orient feature and resized by stretching to a fixed resolution of $640 \times 640$ pixels to ensure uniform input dimensions compatible with the RT-DETR architecture. Data augmentation was applied statically prior to training via the Roboflow platform to increase the dataset's robustness. Specifically, each original training image was used to generate up to three augmented versions by introducing random rotations within a range of $\pm 13^{\circ}$. This specific transformation was selected to increase the model's invariance to small orientation changes while preserving the inherent geometric structure of the spherical objects.

By utilizing this statically augmented, uniformly preprocessed, and filtered dataset across all transformer-based architectures, any observed performance differences during evaluation can be attributed exclusively to architectural variations, dropout regularization, and test illumination conditions, rather than discrepancies in data distribution.

\begin{figure}[ht]
  \centering
  \includegraphics[width=0.8\linewidth]{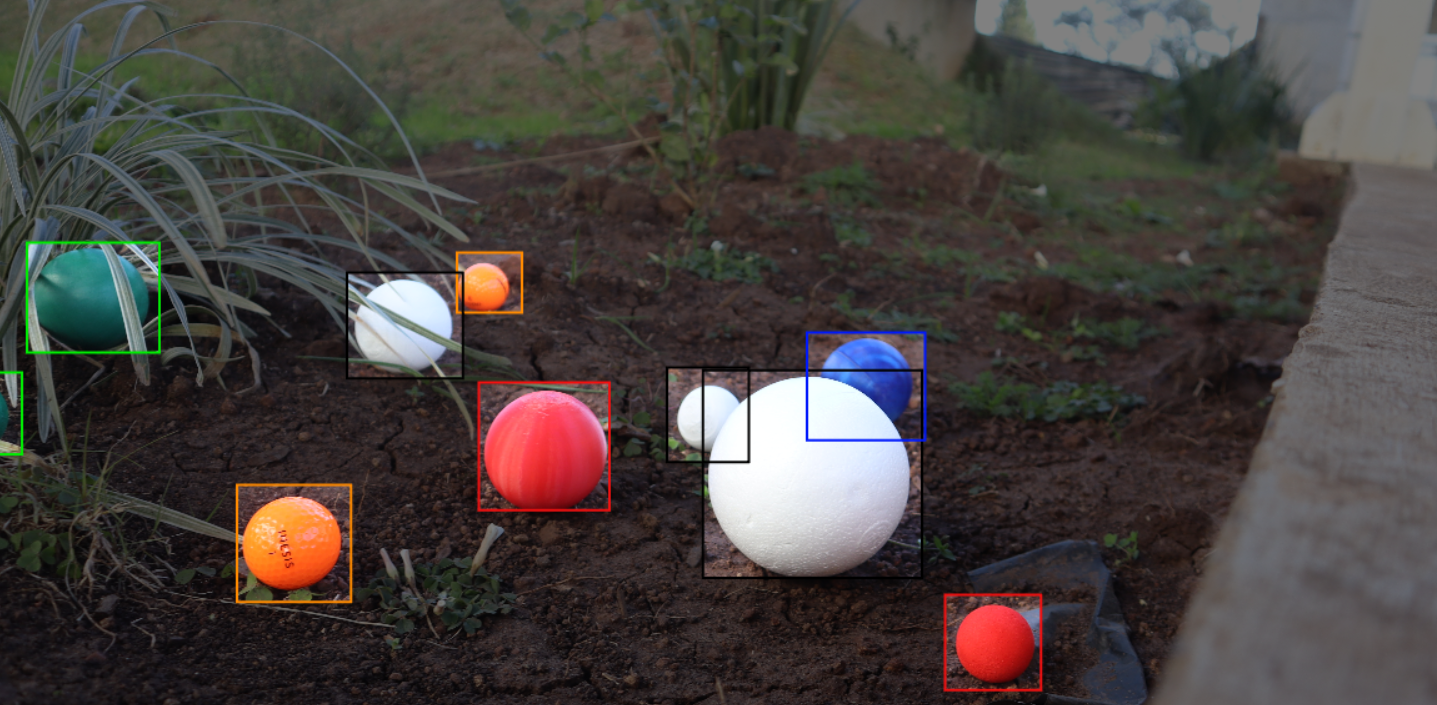}
  \caption{Sample images from the Urubots Ball Detection dataset showing the different object classes and bounding box annotations under varying lighting conditions.}
  \label{fig:Dataset}
\end{figure}

\subsection{Trainig Procedure}
To evaluate the performance of the RT-DETR architecture under varying environmental conditions, all training and inference experiments were conducted on a system equipped with an NVIDIA RTX 4070 Ti GPU. Four variants of the ResNet family were implemented as feature extraction backbones: ResNet18, ResNet34, ResNet50, and ResNet101. To assess the impact of regularization on detection confidence and latency, each architecture was trained in two modes: \textbf{Standard} (without dropout) and \textbf{Regularized} (with a dropout rate of 0.2).

All models were trained from scratch (\texttt{pretrained=False}) for 150 epochs under identical conditions to ensure a fair comparison. The training setup was standardized with an input image size of $640 \times 640$ and a batch size of 4. Since the Ultralytics framework was executed with \texttt{optimizer=auto}, the optimizer hyperparameters were selected automatically during training. In the reported experiments, this resulted in the use of Stochastic Gradient Descent (SGD) with a learning rate of 0.01 and momentum of 0.9. For optimization, the RT-DETR implementation relies on \texttt{RTDETRDetectionLoss}, which extends the DETR loss formulation by combining classification, bounding box, GIoU, and auxiliary losses, with an additional denoising loss when enabled.

\subsection{Evaluation Procedure}
The experiments utilized a controlled dataset composed of spherical objects of different colors (blue, green, orange, and white). After training, evaluation under simulated deployment conditions was performed using videos recorded with a Logitech C270 camera. These test scenarios encompassed four distinct experimental conditions: artificial illumination (Full Light) and reduced illumination (Low Light), each cross-evaluated over white and black backgrounds. This configuration enabled a controlled analysis of how lighting conditions and environmental contrast affect detector performance.

\subsection{Performance Metrics}

Keeping the hardware, hyperparameters, and experimental protocol fixed ensured that any performance differences could be attributed only to backbone depth, dropout regularization, and the evaluated environmental conditions. Under this controlled setup, three complementary metrics were used during evaluation. The first was \textbf{Average Confidence}, defined as the mean confidence score assigned by the detector to the predicted class, which provides an estimate of how certain the model is when producing a detection. The second was \textbf{Inference Time}, measured in milliseconds per frame, which quantifies the temporal cost of the detector and is particularly relevant for real time robotic applications. The third was \textbf{Accuracy by Ball Color}, computed separately for each object class, in order to assess whether the detector preserved consistent classification performance across different hues and under changes in illumination or background conditions. Together, these metrics make it possible to analyze not only whether the model predicts correctly, but also how reliable and how efficient those predictions are in practice.

\section{\uppercase{EXPERIMENTAL RESULTS}}

\noindent The aggregated results across all backbone configurations provide a systematic comparison of architectural depth, regularization, illumination, and background conditions. Performance is reported for both training modes—with a Dropout rate of 0.2 and without Dropout (0.0)—enabling a direct assessment of how regularization impacts detection latency and confidence.

\begin{figure}[th]
  \centering
  \includegraphics[width=0.75\linewidth]{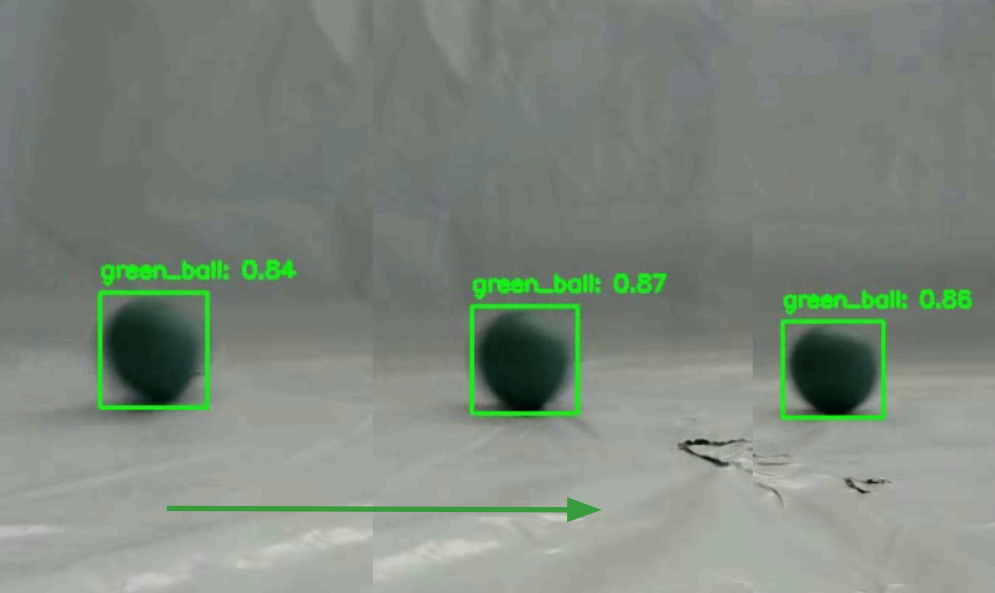}
  \caption{Sequential frames from a real-time inference test, demonstrating the model's ability to detect and track the spherical objects in motion.}\label{fig:Inference_Frames}
\end{figure}

Video sequences of moving balls were used for inference in order to evaluate practical behavior beyond static evaluation. These sequences were only used for performance validation and were captured following model training. The real-time detection procedure is shown in Figure~\ref{fig:Inference_Frames}, where the model locates and follows the target over a series of frames. The bounding box allows for qualitative evaluation of temporal consistency under various dropout and environmental configurations because it maintains spatial stability in the face of object motion.

\subsection{Impact of Dropout under Varying Lighting Conditions}

To examine the interaction between regularization and illumination, a controlled comparison was conducted between the standard configuration with Dropout 0.0 and the regularized configuration with Dropout 0.2, while keeping the backbone architectures, backgrounds, and lighting conditions unchanged. The results are presented in terms of average detection confidence, average inference latency, and classification accuracy.

Tables~\ref{tab:metrics_light_2} and~\ref{tab:metrics_light} present the average detection confidence obtained under \textbf{Full Light} and \textbf{Low Light} conditions for the evaluated backbone architectures and ball colors.

\begin{table}[h!]
\centering
\caption{Average detection confidence under varying illumination conditions (\textbf{Full Light} and \textbf{Low Light}) using the standard model (\textbf{Dropout 0.0}).}
\label{tab:metrics_light_2}
\resizebox{\columnwidth}{!}{
\begin{tabular}{|l|l|c|c|c|c|}
\hline
\multirow{2}{*}{\textbf{Architecture}} & \multirow{2}{*}{\textbf{Lighting}} & \multicolumn{4}{c|}{\textbf{Ball Color}} \\ \cline{3-6}
 & & \textbf{Blue} & \textbf{Green} & \textbf{Orange} & \textbf{White} \\ \hline
\multirow{2}{*}{ResNet18} & Full Light & 0.8913 & 0.8898 & 0.8531 & 0.7627 \\ \cline{2-6}
 & Low Light & 0.8712 & 0.8636 & 0.8247 & 0.7013 \\ \hline
\multirow{2}{*}{ResNet34} & Full Light & 0.6051 & 0.8762 & 0.8661 & 0.8399 \\ \cline{2-6}
 & Low Light & 0.8665 & 0.8779 & 0.8556 & 0.8550 \\ \hline
\multirow{2}{*}{ResNet50} & Full Light & 0.8670 & 0.8707 & 0.8550 & 0.8693 \\ \cline{2-6}
 & Low Light & 0.8388 & 0.8683 & 0.7339 & 0.8610 \\ \hline
\multirow{2}{*}{ResNet101} & Full Light & 0.8612 & 0.8506 & 0.8621 & 0.7144 \\ \cline{2-6}
 & Low Light & 0.8460 & 0.8440 & 0.8111 & 0.7817 \\ \hline
\end{tabular}%
}
\end{table}

\begin{table}[h]
\centering
\caption{Average detection confidence under varying illumination conditions (\textbf{Full Light} and \textbf{Low Light}) with regularization applied (\textbf{Dropout 0.2}).}
\label{tab:metrics_light}
\resizebox{\columnwidth}{!}{%
\begin{tabular}{|l|l|c|c|c|c|}
\hline
\multirow{2}{*}{\textbf{Architecture}} & \multirow{2}{*}{\textbf{Lighting}} & \multicolumn{4}{c|}{\textbf{Ball Color}} \\ \cline{3-6}
 & & \textbf{Blue} & \textbf{Green} & \textbf{Orange} & \textbf{White} \\ \hline
\multirow{2}{*}{ResNet18} & Full Light & 0.8190 & 0.8450 & 0.8051 & 0.8028 \\ \cline{2-6}
 & Low Light & 0.8051 & 0.8031 & 0.7821 & 0.7419 \\ \hline
\multirow{2}{*}{ResNet34} & Full Light & 0.8095 & 0.8321 & 0.8237 & 0.8129 \\ \cline{2-6}
 & Low Light & 0.8341 & 0.8504 & 0.8195 & 0.8524 \\ \hline
\multirow{2}{*}{ResNet50} & Full Light & 0.8316 & 0.8126 & 0.7698 & 0.7078 \\ \cline{2-6}
 & Low Light & 0.7593 & 0.7987 & 0.8101 & 0.6718 \\ \hline
\multirow{2}{*}{ResNet101} & Full Light & 0.7980 & 0.8076 & 0.7797 & 0.8079 \\ \cline{2-6}
 & Low Light & 0.7786 & 0.8248 & 0.7626 & 0.7755 \\ \hline
\end{tabular}%
}
\end{table}

Tables~\ref{tab:detection_light} and~\ref{tab:detection_light_2} present the classification accuracy obtained under varying illumination conditions for Dropout 0.0 and Dropout 0.2, respectively.

\begin{table}[h!]
\centering
\caption{Classification accuracy under varying illumination conditions (\textbf{Full Light} and \textbf{Low Light}) using the standard model (\textbf{Dropout 0.0}).}
\label{tab:detection_light}
\resizebox{\columnwidth}{!}{%
\begin{tabular}{|l|l|c|c|c|c|}
\hline
\multirow{2}{*}{\textbf{Architecture}} & \multirow{2}{*}{\textbf{Lighting}} & \multicolumn{4}{c|}{\textbf{Ball Color}} \\ \cline{3-6}
 & & \textbf{Blue} & \textbf{Green} & \textbf{Orange} & \textbf{White} \\ \hline
\multirow{2}{*}{ResNet18} & Full Light & 17/17 (0) & 23/23 (0) & 20/20 (0) & 13/17 (4) \\ \cline{2-6}
 & Low Light & 19/19 (0) & 47/47 (0) & 26/26 (0) & 31/32 (1) \\ \hline
\multirow{2}{*}{ResNet34} & Full Light & 17/49 (32) & 24/24 (0) & 20/20 (0) & 19/19 (0) \\ \cline{2-6}
 & Low Light & 20/20 (0) & 47/47 (0) & 26/26 (0) & 33/33 (0) \\ \hline
\multirow{2}{*}{ResNet50} & Full Light & 18/18 (0) & 24/24 (0) & 20/20 (0) & 17/17 (0) \\ \cline{2-6}
 & Low Light & 20/20 (0) & 46/46 (0) & 27/27 (0) & 31/31 (0) \\ \hline
\multirow{2}{*}{ResNet101} & Full Light & 18/18 (0) & 23/23 (0) & 20/20 (0) & 18/20 (2) \\ \cline{2-6}
 & Low Light & 19/19 (0) & 47/47 (0) & 26/26 (0) & 29/29 (0) \\ \hline
\end{tabular}%
}
\end{table}

\begin{table}[h!]
\centering
\caption{Classification accuracy under varying illumination conditions (\textbf{Full Light} and \textbf{Low Light}) with regularization applied (\textbf{Dropout 0.2}).}
\label{tab:detection_light_2}
\resizebox{\columnwidth}{!}{%
\begin{tabular}{|l|l|c|c|c|c|}
\hline
\multirow{2}{*}{\textbf{Architecture}} & \multirow{2}{*}{\textbf{Lighting}} & \multicolumn{4}{c|}{\textbf{Ball Color}} \\ \cline{3-6}
 & & \textbf{Blue} & \textbf{Green} & \textbf{Orange} & \textbf{White} \\ \hline
\multirow{2}{*}{ResNet18} & Full Light & 33/34 (1) & 50/50 (0) & 42/42 (0) & 40/41 (1) \\ \cline{2-6}
 & Low Light & 45/45 (0) & 69/69 (0) & 48/48 (0) & 56/56 (0) \\ \hline
\multirow{2}{*}{ResNet34} & Full Light & 39/39 (0) & 48/48 (0) & 40/40 (0) & 39/39 (0) \\ \cline{2-6}
 & Low Light & 44/44 (0) & 66/66 (0) & 50/50 (0) & 32/32 (0) \\ \hline
\multirow{2}{*}{ResNet50} & Full Light & 23/24 (1) & 45/45 (0) & 41/41 (0) & 30/32 (2) \\ \cline{2-6}
 & Low Light & 46/46 (0) & 72/72 (0) & 49/49 (0) & 25/45 (20) \\ \hline
\multirow{2}{*}{ResNet101} & Full Light & 27/27 (0) & 50/50 (0) & 45/45 (0) & 36/37 (1) \\ \cline{2-6}
 & Low Light & 46/46 (0) & 69/69 (0) & 52/52 (0) & 55/55 (0) \\ \hline
\end{tabular}%
}
\end{table}

\begin{figure}[h!]
  \centering
  \includegraphics[width=0.8\linewidth]{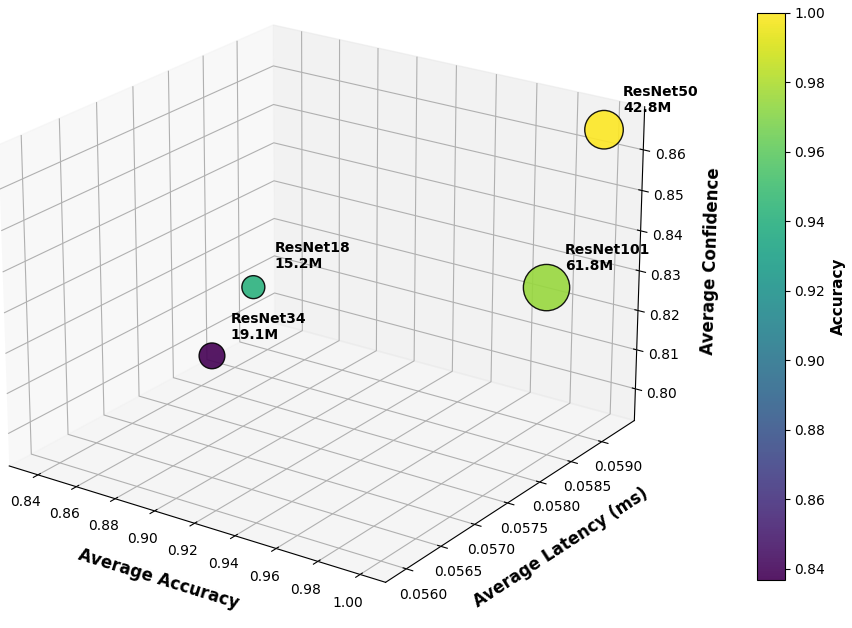}
  \caption{Trade-off among average accuracy, average latency, average confidence, and model size for the best-performing condition in the illumination variation experiment (\textbf{Full Light}, \textbf{Dropout 0.0}). Each point represents a backbone architecture, and the marker size is proportional to the total number of model parameters.}
  \label{fig:graf1}
\end{figure}

Figure~\ref{fig:graf1} summarizes the results obtained for the best-performing condition in the illumination variation experiment, combining average accuracy, average latency, average confidence, and model size for each evaluated backbone architecture.

\subsection{Impact of Dropout and Background}

To evaluate the effect of regularization under background variation, models trained with Dropout 0.2 were compared with non-regularized models (Dropout 0.0) under white and black backgrounds. Since background color changes object contrast, this analysis was used to measure its effect on detection confidence, accuracy, and inference latency. All other training hyperparameters and evaluation settings were kept fixed, so the observed differences can be attributed to dropout and background contrast.

Tables~\ref{tab:metrics_bg} and~\ref{tab:metrics_bg_2} present the average confidence under constant illumination conditions (Full Light), comparing configurations with Dropout 0.0 and Dropout 0.2 on white and black backgrounds.
\begin{table}[h]
\centering
\caption{Average detection confidence under constant illumination (\textbf{Full Light}) for white and black background conditions using the standard model (\textbf{Dropout 0.0}).}
\label{tab:metrics_bg}
\resizebox{\columnwidth}{!}{%
\begin{tabular}{|l|l|c|c|c|c|}
\hline
\multirow{2}{*}{\textbf{Architecture}} & \multirow{2}{*}{\textbf{Background}} & \multicolumn{4}{c|}{\textbf{Ball Color}} \\ \cline{3-6} 
 & & \textbf{Blue} & \textbf{Green} & \textbf{Orange} & \textbf{White} \\ \hline
\multirow{2}{*}{ResNet18} & Black BG & 0.7558 & 0.8546 & 0.7839 & 0.6485 \\ \cline{2-6} 
 & White BG & 0.8484 & 0.8898 & 0.8531 & 0.7962 \\ \hline
\multirow{2}{*}{ResNet34} & Black BG & 0.7698 & 0.8438 & 0.7989 & 0.7742 \\ \cline{2-6} 
 & White BG & 0.6051 & 0.8762 & 0.8661 & 0.8399 \\ \hline
\multirow{2}{*}{ResNet50} & Black BG & 0.6507 & 0.7202 & 0.5531 & 0.7548 \\ \cline{2-6} 
 & White BG & 0.8670 & 0.8707 & 0.8550 & 0.8693 \\ \hline
\multirow{2}{*}{ResNet101} & Black BG & 0.7133 & 0.7023 & 0.6972 & 0.7726 \\ \cline{2-6} 
 & White BG & 0.8612 & 0.8506 & 0.8621 & 0.7144 \\ \hline
\end{tabular}%
}
\end{table}

\begin{table}[th!]
\centering
\caption{Average detection confidence under constant illumination (\textbf{Full Light}) for white and black background conditions with regularization applied (\textbf{Dropout 0.2}).}
\label{tab:metrics_bg_2}
\resizebox{\columnwidth}{!}{%
\begin{tabular}{|l|l|c|c|c|c|}
\hline
\multirow{2}{*}{\textbf{Architecture}} & \multirow{2}{*}{\textbf{Background}} & \multicolumn{4}{c|}{\textbf{Ball Color}} \\ \cline{3-6} 
 & & \textbf{Blue} & \textbf{Green} & \textbf{Orange} & \textbf{White} \\ \hline
\multirow{2}{*}{ResNet18} & Black BG & 0.7658 & 0.8185 & 0.7706 & 0.8112 \\ \cline{2-6} 
 & White BG & 0.8663 & 0.8738 & 0.8431 & 0.7931 \\ \hline
\multirow{2}{*}{ResNet34} & Black BG & 0.7493 & 0.7926 & 0.7808 & 0.7624 \\ \cline{2-6} 
 & White BG & 0.8874 & 0.8715 & 0.8665 & 0.8660 \\ \hline
\multirow{2}{*}{ResNet50} & Black BG & 0.7119 & 0.7780 & 0.6903 & 0.5716 \\ \cline{2-6} 
 & White BG & 0.8715 & 0.8379 & 0.8455 & 0.8138 \\ \hline
\multirow{2}{*}{ResNet101} & Black BG & 0.6395 & 0.7729 & 0.7217 & 0.7796 \\ \cline{2-6} 
 & White BG & 0.8647 & 0.8396 & 0.8459 & 0.8348 \\ \hline
\end{tabular}%
}
\end{table}

In Tables~\ref{tab:detection_bg} and~\ref{tab:detection_bg_2}, the average accuracy results under constant illumination conditions (Full Light) are presented, comparing configurations with Dropout 0.2 and Dropout 0.0 over white and black backgrounds. When the average detections across the evaluated video contained no classification errors, the accuracy metric was set to 1; otherwise, when misclassifications occurred, the values were lower.

\begin{table}[h!]
\centering
\caption{Classification accuracy under constant illumination (\textbf{Full Light}) for white and black background conditions using the standard model (\textbf{Dropout 0.0}).}
\label{tab:detection_bg}
\resizebox{\columnwidth}{!}{%
\begin{tabular}{|l|l|c|c|c|c|}
\hline
\multirow{2}{*}{\textbf{Architecture}} & \multirow{2}{*}{\textbf{Background}} & \multicolumn{4}{c|}{\textbf{Ball Color}} \\ \cline{3-6} 
 & & \textbf{Blue} & \textbf{Green} & \textbf{Orange} & \textbf{White} \\ \hline
\multirow{2}{*}{ResNet18} & Black BG & 22/22 (0) & 24/24 (0) & 20/20 (0) & 18/18 (0) \\ \cline{2-6} 
 & White BG & 34/51 (17) & 46/46 (0) & 20/20 (0) & 26/54 (28) \\ \hline
\multirow{2}{*}{ResNet34} & Black BG & 25/25 (0) & 24/24 (0) & 22/22 (0) & 22/22 (0) \\ \cline{2-6} 
 & White BG & 34/98 (64) & 48/48 (0) & 40/40 (0) & 38/38 (0) \\ \hline
\multirow{2}{*}{ResNet50} & Black BG & 8/8 (0) & 22/22 (0) & 12/12 (0) & 18/18 (0) \\ \cline{2-6} 
 & White BG & 36/36 (0) & 48/48 (0) & 40/40 (0) & 34/34 (0) \\ \hline
\multirow{2}{*}{ResNet101} & Black BG & 20/20 (0) & 26/26 (0) & 22/22 (0) & 16/16 (0) \\ \cline{2-6} 
 & White BG & 36/36 (0) & 46/46 (0) & 40/40 (0) & 36/40 (4) \\ \hline
\end{tabular}%
}
\end{table}

\begin{table}[h!]
\centering
\caption{Classification accuracy under constant illumination (\textbf{Full Light}) for white and black background conditions with regularization applied (\textbf{Dropout 0.2}).}
\label{tab:detection_bg_2}
\resizebox{\columnwidth}{!}{%
\begin{tabular}{|l|l|c|c|c|c|}
\hline
\multirow{2}{*}{\textbf{Architecture}} & \multirow{2}{*}{\textbf{Background}} & \multicolumn{4}{c|}{\textbf{Ball Color}} \\ \cline{3-6} 
 & & \textbf{Blue} & \textbf{Green} & \textbf{Orange} & \textbf{White} \\ \hline
\multirow{2}{*}{ResNet18} & Black BG & 32/32 (0) & 52/52 (0) & 44/44 (0) & 44/44 (0) \\ \cline{2-6} 
 & White BG & 17/18 (1) & 48/48 (0) & 40/40 (0) & 36/38 (2) \\ \hline
\multirow{2}{*}{ResNet34} & Black BG & 44/44 (0) & 48/48 (0) & 40/40 (0) & 40/40 (0) \\ \cline{2-6} 
 & White BG & 34/34 (0) & 48/48 (0) & 40/40 (0) & 38/38 (0) \\ \hline
\multirow{2}{*}{ResNet50} & Black BG & 12/12 (0) & 38/38 (0) & 40/40 (0) & 28/28 (0) \\ \cline{2-6} 
 & White BG & 34/36 (2) & 52/52 (0) & 42/42 (0) & 32/36 (4) \\ \hline
\multirow{2}{*}{ResNet101} & Black BG & 16/16 (0) & 48/48 (0) & 48/48 (0) & 36/36 (0) \\ \cline{2-6} 
 & White BG & 38/38 (0) & 52/52 (0) & 42/42 (0) & 36/38 (2) \\ \hline
\end{tabular}%
}
\end{table}

\begin{figure}[ht!]
  \centering
  \includegraphics[width=0.8\linewidth]{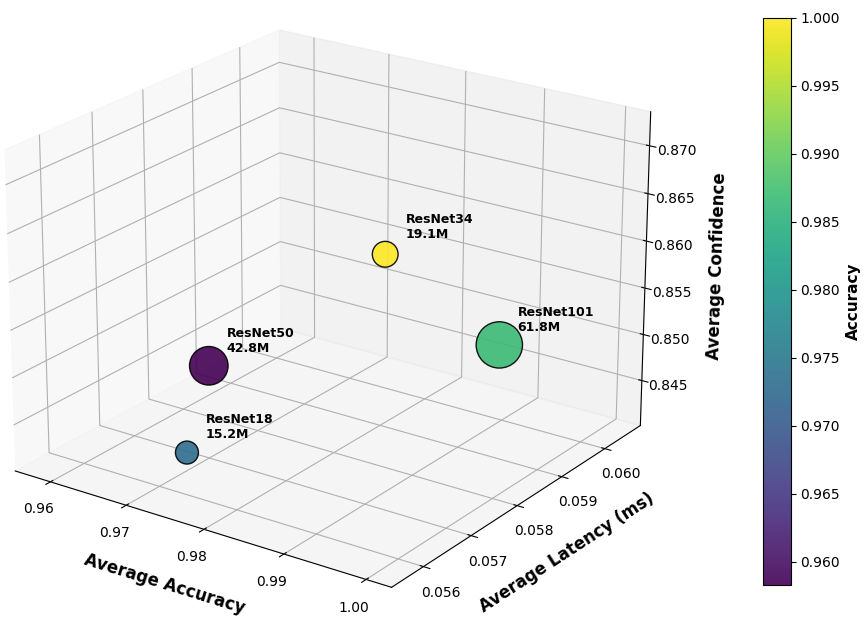}
  \caption{Trade off among average accuracy, average latency, average confidence, and model size for the best performing condition in the background variation experiment (\textbf{White Background}, \textbf{Full Light}, \textbf{Dropout 0.2}). Each point represents a backbone architecture, and the marker size is proportional to the total number of model parameters.}\label{fig:graf2}
\end{figure}

Figure~\ref{fig:graf2} highlights the best result obtained in the background variation experiment, corresponding to the \textit{White Background}, \textit{Full Light}, and \textbf{Dropout 0.2} condition.

\section{\uppercase{Discussion}}

The results obtained under varying lighting conditions provide a comprehensive view of how illumination affects the behavior of the detector when comparing the standard configuration (\textbf{Dropout 0.0}) with the regularized version (\textbf{Dropout 0.2}). By evaluating both \textit{Full Light} and \textit{Low Light} scenarios while keeping all other factors constant, it is possible to isolate the effect of illumination on different performance indicators. Overall, the results show that lighting variation impacts prediction confidence more strongly than latency, while classification accuracy remains high in most cases.

Tables~\ref{tab:metrics_light_2} and~\ref{tab:metrics_light} present the average detection confidence under \textit{Full Light} and \textit{Low Light} conditions for the two regularization settings. Compared with accuracy, confidence reveals a higher sensitivity to illumination changes. Under \textbf{Dropout 0.0}, confidence values show greater variability when transitioning from \textit{Full Light} to \textit{Low Light}, with noticeable reductions in several architectures and classes, particularly for colors such as white and orange. With \textbf{Dropout 0.2}, confidence values become more stable in many cases, and the gap between lighting conditions is reduced. However, this improvement is not uniform across all backbones, suggesting that the effect of dropout depends on both the architecture and the visual characteristics of the object.

Tables~\ref{tab:detection_light} and~\ref{tab:detection_light_2} show the classification accuracy obtained under varying illumination conditions for Dropout 0.0 and Dropout 0.2, respectively. In general, both configurations achieve consistently strong results across most backbone architectures and object classes. Under \textbf{Dropout 0.0}, only a few specific cases show noticeable degradation, particularly in certain class--architecture combinations such as the blue class with ResNet34 and the white class in some deeper models. When \textbf{Dropout 0.2} is introduced, accuracy tends to become more uniform across conditions, reducing some of these inconsistencies. However, isolated failures still appear, indicating that while dropout improves stability, it does not completely eliminate sensitivity to challenging lighting conditions.


When comparing backbone architectures, the results indicate that model depth alone does not guarantee better robustness under illumination changes. While deeper models such as \textbf{ResNet50} and \textbf{ResNet101} achieve high accuracy, their behavior differs in terms of efficiency and confidence. \textbf{ResNet50} consistently provides the highest confidence values while maintaining competitive latency, making it the strongest performer overall. In contrast, although \textbf{ResNet101} has a larger model capacity, it does not provide a clear improvement over ResNet50, indicating diminishing returns with increasing depth. Lighter models such as \textbf{ResNet18} offer efficient solutions with lower computational cost, although with slightly reduced confidence under more challenging conditions. Intermediate architectures such as \textbf{ResNet34} provide stable results, but do not consistently outperform either lighter or deeper models.

Figure~\ref{fig:graf1} summarizes the best operating point obtained in the illumination variation experiment, combining average accuracy, average latency, average confidence, and model size for each backbone architecture. This representation highlights the trade-off between predictive performance and model complexity.

The Figure~\ref{fig:graf1} shows that \textbf{ResNet50} achieves the best overall performance, reaching the highest combination of average accuracy and average confidence while maintaining latency comparable to the other architectures. This positions ResNet50 as the most effective model under the evaluated conditions. Additionally, the results confirm that the relationship between model size and performance is not strictly monotonic, as \textbf{ResNet101} does not outperform ResNet50 despite its larger size.

Overall, the analysis suggests that illumination primarily affects the reliability of predictions rather than their correctness or computational cost. While dropout improves stability in several cases, its impact depends on the architecture and does not fully compensate for challenging lighting conditions.

The results under background variation provide a complementary view of the detector behavior under controlled illumination, allowing the effect of object--background contrast to be isolated more clearly. In this setting, the comparison between white and black backgrounds shows that background changes do not affect all performance indicators equally. While accuracy remains high in most cases and latency is largely unchanged, confidence is more sensitive to contrast variation, revealing differences between backbone architectures and regularization schemes that are not always visible from accuracy alone.

Tables~\ref{tab:detection_bg} and~\ref{tab:detection_bg_2} show the average accuracy obtained under \textit{Full Light} for the two regularization settings, Dropout 0.0 and Dropout 0.2, respectively. In general, most architecture--background combinations achieved an accuracy of 1.0, indicating stable classification performance under controlled conditions. However, class specific differences can still be observed. Without dropout, the most evident reductions appear in the blue and white classes, especially when object--background contrast is less favorable. Once Dropout 0.2 is introduced, these errors are reduced or disappear, and the accuracy becomes more uniform across architectures and backgrounds. This suggests that dropout improves classification stability when the appearance of the scene changes due to contrast differences.

A more detailed effect of background variation is observed in the confidence results presented in Tables~\ref{tab:metrics_bg} and~\ref{tab:metrics_bg_2}. Unlike accuracy, which only reflects whether the final prediction is correct, confidence indicates how certain the detector is when assigning the class label. Under Dropout 0.0, confidence values are more dispersed across both architectures and backgrounds, with the black background producing larger reductions in several cases. This behavior indicates that, even when the detector still classifies correctly, the internal representation becomes less stable under contrast changes. With Dropout 0.2, the confidence values become more homogeneous, and the gap between white and black backgrounds is reduced. This effect is especially noticeable in classes such as blue, orange, and white, which are more sensitive to contrast differences. Therefore, the confidence analysis suggests that dropout does not simply reduce classification errors but also improves the robustness of the learned representation under background variation.


Finally, Figure~\ref{fig:graf2} summarizes the best operating point obtained in the background variation experiment, corresponding to the \textit{White Background} condition under \textit{Full Light} with \textbf{Dropout 0.2}. In this representation, each backbone is described by its average accuracy, average latency, and average confidence, while the bubble size reflects the number of model parameters. This makes it possible to compare not only predictive performance but also the trade off between robustness and model complexity under the most favorable background condition identified in the experiments.

The Figure~\ref{fig:graf2} shows that the four architectures do not follow a simple monotonic relationship between size and performance. \textbf{ResNet34} provides the most favorable overall result in this scenario, since it combines perfect average accuracy with the highest average confidence while maintaining low latency and a relatively compact model size. In practical terms, this places ResNet34 at the most balanced point of the trade off, offering the best compromise between detection reliability and computational cost.

\section{\uppercase{Conclusions}}

RT-DETR was evaluated with ResNet18, ResNet34, ResNet50, and ResNet101 under controlled changes in illumination, background contrast, and dropout regularization. The experimental results demonstrate that illumination and background variation have a greater impact on prediction reliability than on accuracy or computational cost; classification accuracy is high in most configurations, but confidence is more sensitive to changes in lighting and contrast, suggesting that internal feature representations are impacted even when predictions are accurate.

Dropout reduces variability in accuracy and confidence across conditions and improves stability in a number of scenarios. However, its impact varies depending on the data's visual qualities and the backbone architecture. Dropout, in particular, lessens sensitivity to changes in illumination and background, but in more difficult situations, it does not totally prevent performance degradation.

The findings show that robustness is not always improved by increasing model depth across architectures. While ResNet34 offers the most balanced performance under background variation, ResNet50 achieves the best overall trade-off between accuracy, confidence, and efficiency under illumination variation. These results imply that, under real-world circumstances, intermediate architectures may provide more stable behavior than deeper models.

Overall, the study shows that while no single configuration can completely compensate for difficult visual conditions, architectural decisions and regularization play a significant role in enhancing robustness, and that environmental factors like lighting and contrast have a greater impact on model confidence than accuracy.

\section*{\uppercase{Acknowledgements }}
We would like to express our sincere gratitude to the Department of Robotics and IA for all the support in hardware to develop this investigation. 

The authors recognize the use of tools like GPT-5 and Gemine Pro for grammar revision, orthography correction, and improving the fluency throughout the text to achieve the standards of linguistics in the conference. The technical content and the scientific contributions were completely developed by the authors.

\bibliographystyle{apalike}
{\small
\bibliography{example}}

\end{document}